\documentclass[conference]{IEEEtran}  % Comment this line out if you need a4paper

\IEEEoverridecommandlockouts                              % This command is only needed if
\usepackage{graphicx}
\usepackage[caption=false, font=footnotesize]{subfig}
\usepackage{cite}
\usepackage{textcomp}
\usepackage{svg}
\usepackage{amsmath}
\usepackage{soul}
\usepackage{flushend}
\usepackage{pdfpages}
\usepackage{fancyvrb}

\usepackage{tikz,verbatimbox}
\usetikzlibrary{shapes,arrows,fit, backgrounds,positioning}
\pgfdeclarelayer{bg}    % declare background layer
\pgfsetlayers{bg,main}  % set the order of the layers (main is the standard layer)
\def\checkmark{\tikz\fill[scale=0.3](0,.35) -- (.25,0) -- (1,.7) -- (.25,.15) -- cycle;}

\title{\LARGE \bf
Simulation Framework for Mobile Robots in Planetary-Like Environments
}

\author{\IEEEauthorblockN{Riccardo Giubilato$^{(1,3)(*)}$, Andrea Masili$^{(2)}$, Sebastiano Chiodini$^{(1,2)}$, Marco Pertile$^{(1,2)}$ and Stefano Debei$^{(1,2)}$}% <-this % stops a space
\IEEEauthorblockA{$^{(1)}$ CISAS "G. Colombo'', University of Padova, Via Venezia 15, 35131 Padova, Italy\\
$^{(2)}$ Department of Industrial Engineering, University of Padova, Via Venezia 1, 35131 Padova, Italy\\
$^{(3)}$ DLR, Institute of Robotics and Mechatronics, M{\"u}nchener Str. 20, 82234, Wessling, Germany \\
$^{(*)}$ \textit{corresponding author:} riccardo.giubilato@dlr.de
}}

\newcommand\copyrighttext{%
  \footnotesize © 2020 IEEE. Personal use of this material is permitted.  Permission from IEEE must be obtained for all other uses, in any current or future media, including reprinting/republishing this material for advertising or promotional purposes, creating new collective works, for resale or redistribution to servers or lists, or reuse of any copyrighted component of this work in other works.}
\newcommand\copyrightnotice{%
\begin{tikzpicture}[remember picture,overlay]
\node[anchor=south,yshift=200pt] at (current page.south) {\fbox{\parbox{\dimexpr\textwidth-\fboxsep-\fboxrule\relax}{\copyrighttext}}};
\end{tikzpicture}%
}

\begin{document}
\thispagestyle{empty}
\noindent
Cite this paper as:
\begin{myverbbox}{\bibtexCode}
@INPROCEEDINGS{giubilato2020simulation,
  author    = {R. Giubilato and A. Masili and S. Chiodini and
               M. Pertile and S. Debei},
  booktitle = {2020 IEEE 7th International Workshop on Metrology
               for AeroSpace (MetroAeroSpace)},
  title     = {Simulation Framework for Mobile Robots in
               Planetary-Like Environments},
  year      = {2020},
}
\end{myverbbox}
\begin{tikzpicture}[remember picture,overlay]
\node[anchor=north,yshift=-80pt,fill=black!10,draw,rounded corners] at (current page.north) {\bibtexCode};
\end{tikzpicture}%

\copyrightnotice{}
\bstctlcite{IEEEexample:BSTcontrol}

\maketitle
\thispagestyle{empty}
\pagestyle{empty}

%%%%%%%%%%%%%%%%%%%%%%%%%%%%%%%%%%%%%%%%%%%%%%%%%%%%%%%%%%%%%%%%%%%%%%%%%%%%%%%%
\begin{abstract}
%\st{In this paper we present a simulation framework for a mobile robotic platform based on ROS (Robot Operating System) Gazebo.}
In this paper we present a simulation framework for the evaluation of the navigation and localization metrological performances of a robotic platform. The simulator, based on ROS (Robot Operating System) Gazebo, is targeted to a planetary-like research vehicle which allows to test
%\st{alternative perception and navigation strategies relative to different terrain and illumination conditions}
various perception and navigation approaches for specific environment conditions. The possibility of simulating arbitrary sensor setups comprising cameras, LiDARs (Light Detection and Ranging) and IMUs makes Gazebo an excellent resource for rapid prototyping. In this work we evaluate a variety of open-source visual and LiDAR SLAM (Simultaneous Localization and Mapping) algorithms in a
%\st{different simulated natural environments}
simulated Martian environment. Datasets are captured by driving the rover and recording sensors outputs as well as the ground truth for a precise performance evaluation.
\end{abstract}

\begin{IEEEkeywords}
Simulation, Robotics, Navigation, Robot Operating System, Gazebo
\end{IEEEkeywords}

\section{Introduction}
Designing a mobile robot is a costly task, often carried out in an inevitable trial-and-error process. For this reason, simulation toolkits are precious assets to optimize both time and expenses. Aside from mechanical or physical analysis software, which allow to evaluate in detail very specific design choices, many solutions are available to assist the high-level design of the whole robot such as Gazebo\footnote{gazebosim.org}, V-REP\footnote{coppeliarobotics.com/coppeliaSim} or Microsoft AirSim\footnote{github.com/microsoft/AirSim} \cite{shah2018airsim}. This family of software offers simple physical simulation capabilities in order to allow the robot to interact with a simulated environment and more importantly, provides solutions to simulate the output of various types of perception sensors, such as cameras, range sensors, or Inertial Measurement Units (IMU).

Among them, the Gazebo simulator offers a tight integration within the Robot Operating System (ROS) where the generated sensors outputs can be processed by Simultaneous Localization and Mapping (SLAM) algorithms for pose estimation and mapping \cite{giubilato2020relocalization,chiodini2020retrieving,chiodini2019rover,giubilato2019evaluation}. Then, motion planning algorithms can output motor controls which move the robot in a virtual environment.
This is beneficial not only to assist the design process of the robot but also to test thoroughly and in different operating conditions all the algorithms involved. Furthermore, the tight integration with ROS allows to share the source code for all operations between the real robot and the virtual counterpart. This ensures that, when deployed on the field, the real robot will behave almost exactly as foreseen and tested in simulation.
In addition, the evaluation of positioning algorithms on simulated environments is beneficial from the metrological perspective: the ground truth is \textit{exact}, while during field testing it is indeed affected by errors. Lastly, it is possible to evaluate the impact of sensor characteristics, such as FOV and resolution, on the reconstructed trajectory.

\begin{figure}[t]
    \centering
    \subfloat[]{
       \includegraphics[width=0.45\linewidth, trim=0 0cm 0 0cm, clip]{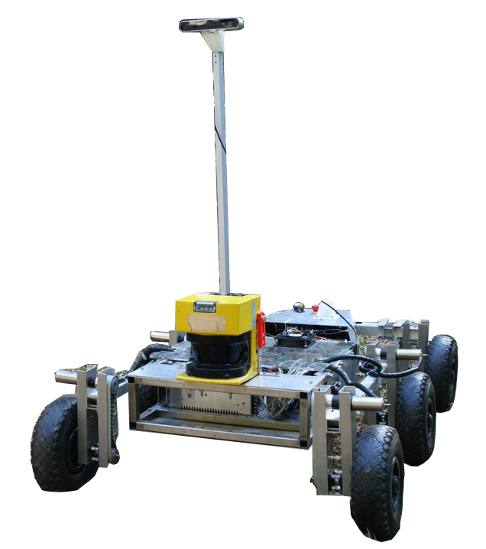}\label{fig::morph_real}} \hfill
    \subfloat[]{%
       \includegraphics[width=0.45\linewidth, trim=0cm 0cm 0cm 0cm, clip]{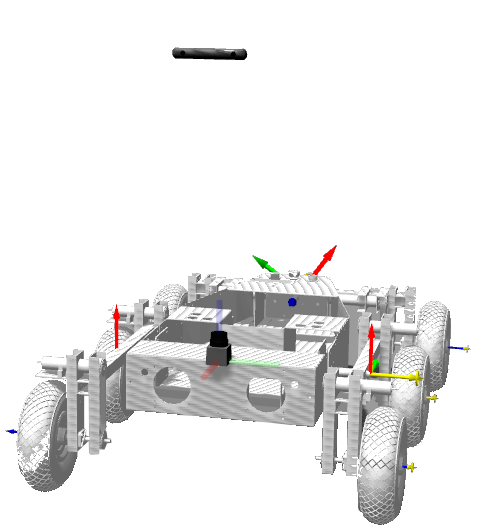}\label{fig::morph_replica}}  \\
    \subfloat[]{%
       \includegraphics[width=1\linewidth]{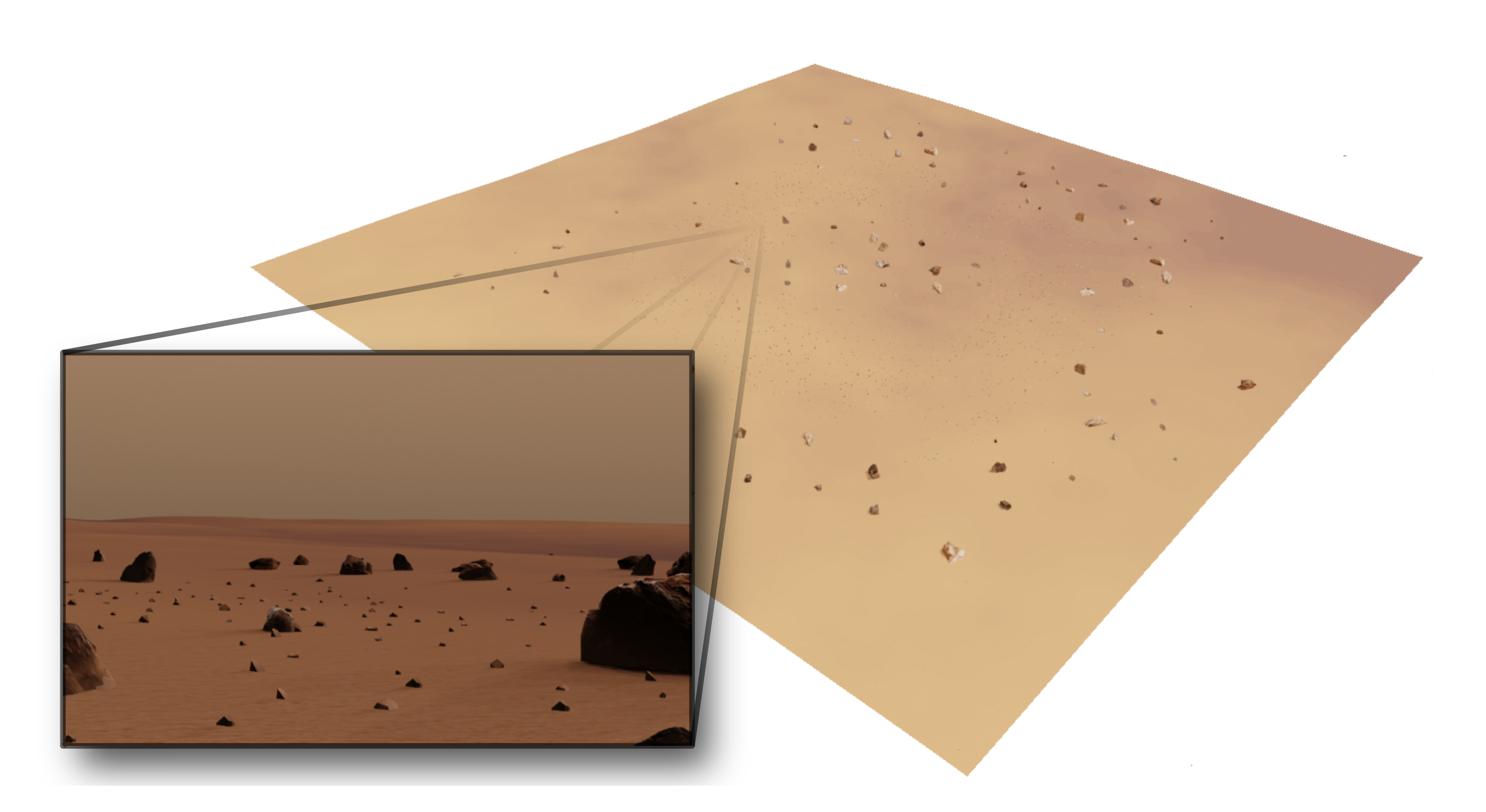}\label{fig::gale}}
    \caption{(a) The MORPHEUS rover \cite{giubilato2019evaluation} and its simulated counterpart in ROS Gazebo. (b) View of a synthetic environment modeled in Blender along with a rendered camera view}
    \label{fig:my_label}
\end{figure}
\begin{figure*}[!htb]
    \centering
    \begin{minipage}[b]{.49\textwidth}
        \subfloat[Left Image]{%
           \includegraphics[width=0.48\linewidth]{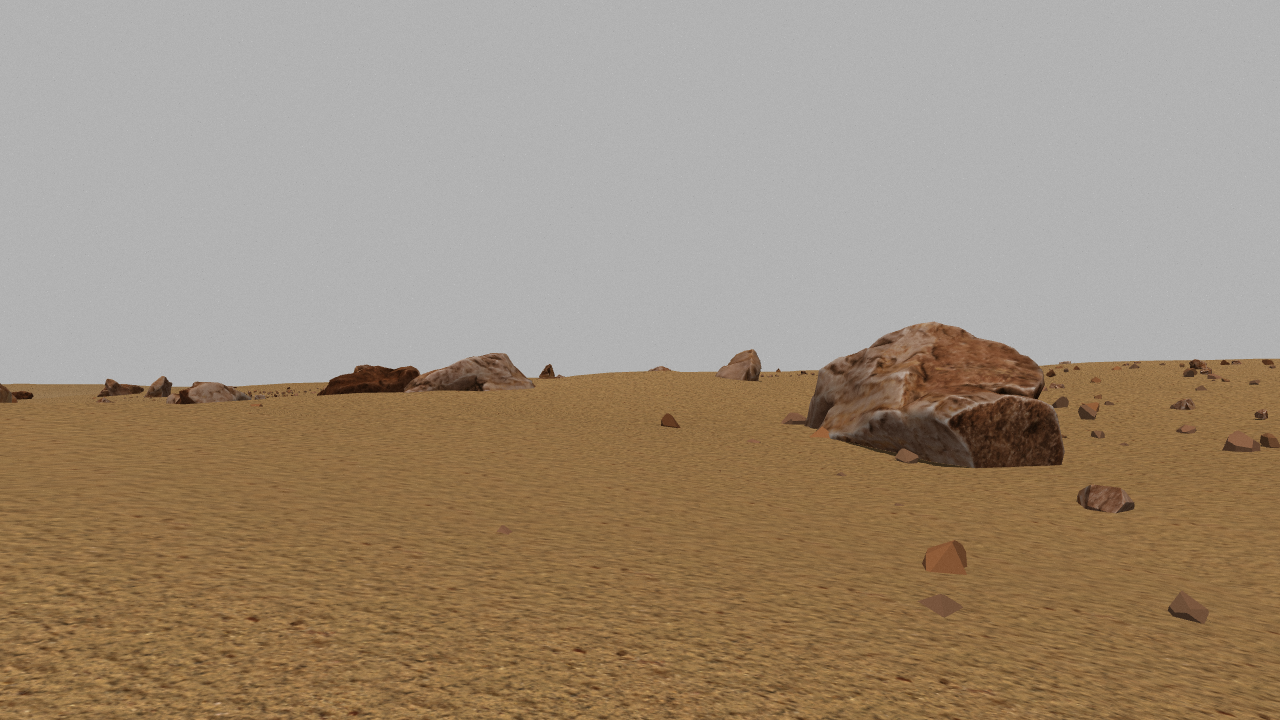}\label{fig::example_left}} \hfill
        \subfloat[Right Image]{%
           \includegraphics[width=0.48\linewidth]{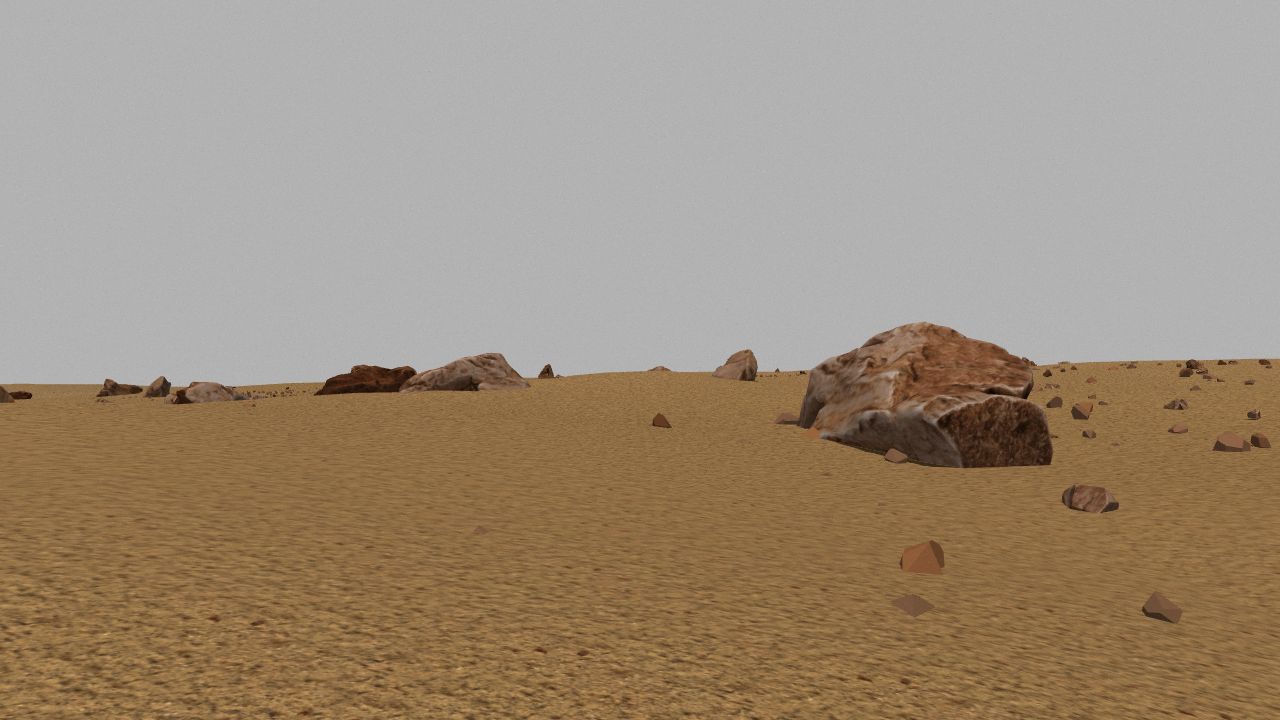}\label{fig::example_right}}  \\
        \subfloat[Disparity]{%
           \includegraphics[width=0.48\linewidth]{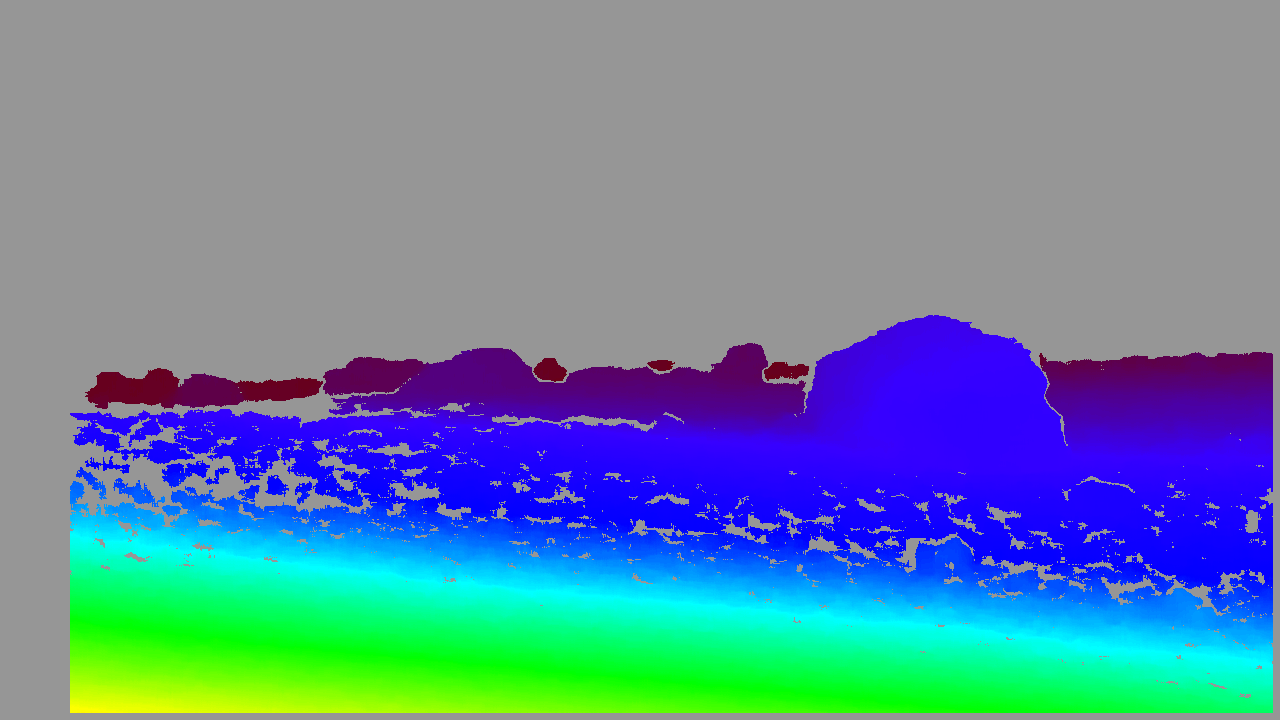}\label{fig::example_disp}} \hfill
        \subfloat[Point Cloud]{%
           \includegraphics[width=0.48\linewidth]{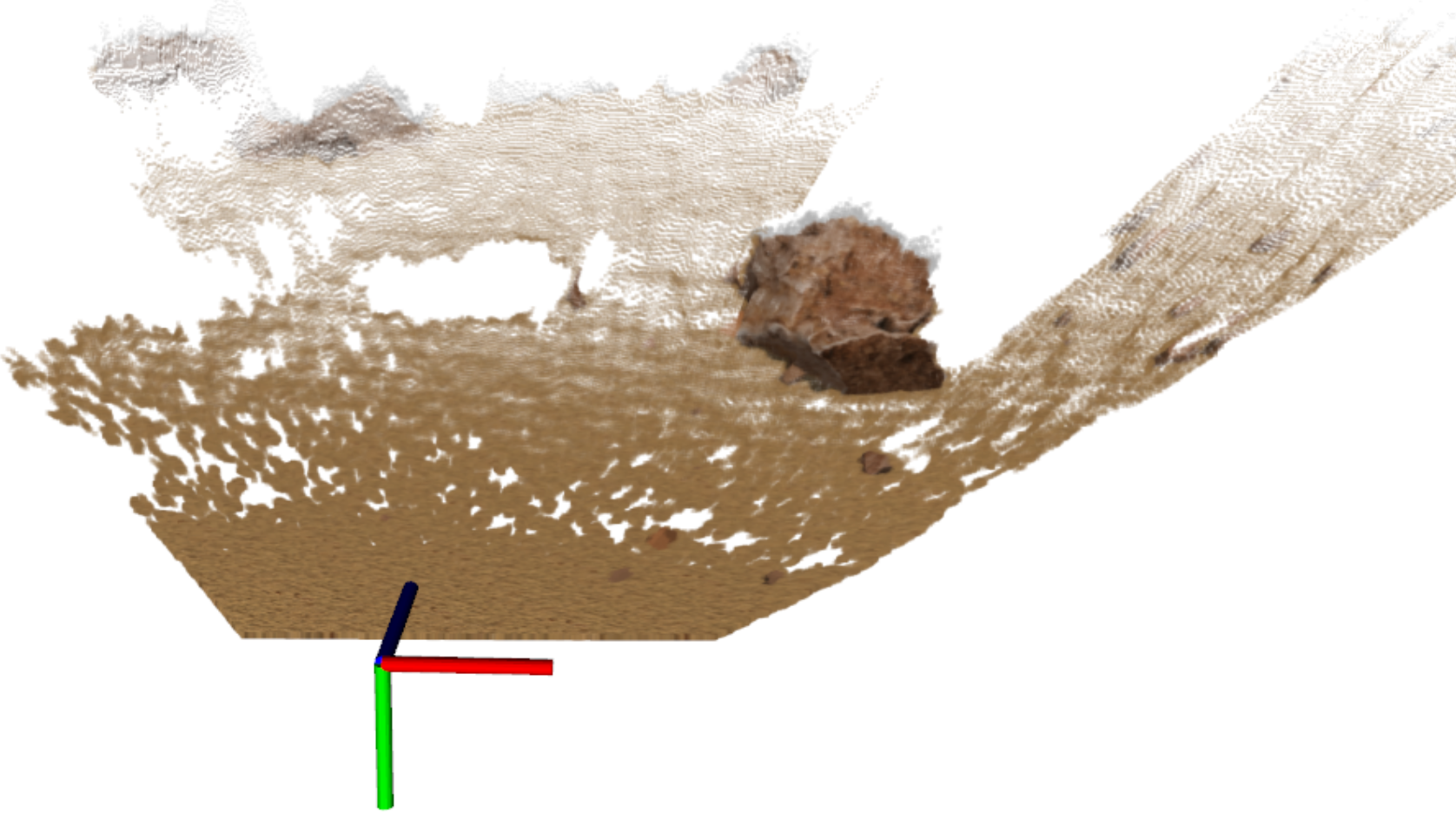}\label{fig::example_cloud}}
    \end{minipage}
    \hfill
    \begin{minipage}[b]{.49\textwidth}
    \subfloat[3D LiDAR scan]{%x
       \includegraphics[width=\linewidth]{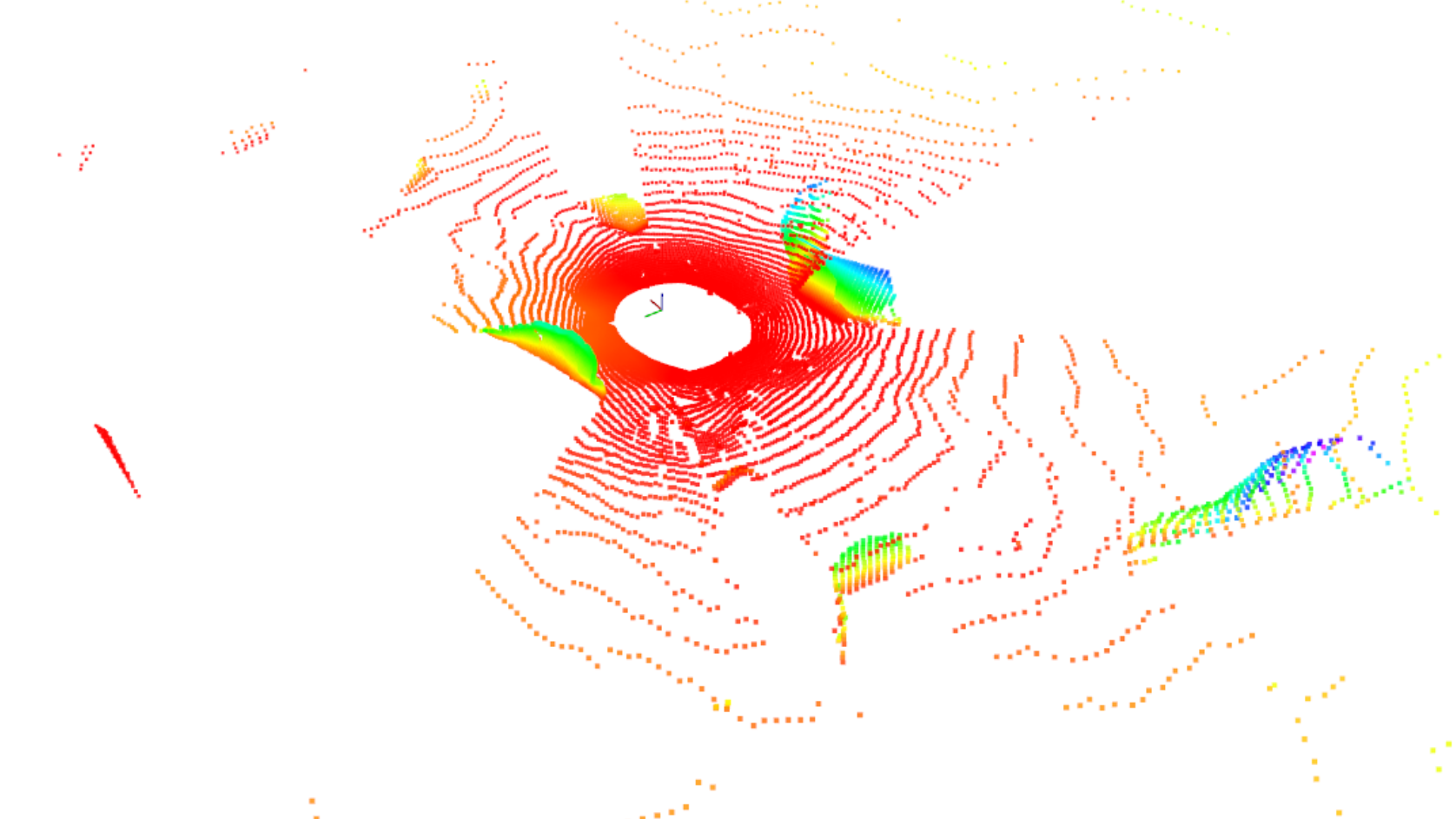}\label{fig::lidar_scan}}
    \end{minipage}
    \caption{Simulated sensing modalities for the MORPHEUS rover. (a-d) Stereo camera output with left and right images (after intrinsic and extrinsic calibration) using the \texttt{multicamera} plugin, disparity map and generated pointcloud. (e) full 3D LiDAR scan generated by the \texttt{gazebo\_ros\_laser\_controller} plugin, colormapped by height}
    \label{fig:my_label}
\end{figure*}
%\SCmod{
In this paper, we present a simulation framework dedicated to the validation of SLAM algorithms given the mobility capabilities of a rover and the Martian topography. The framework is based on ROS and Gazebo, and is targeted to the MORPHEUS rover \cite{giubilato2019evaluation}, a research test-bed for autonomous space operations developed at the University of Padova (see Fig. \ref{fig::morph_real}).
%}
%\st{In this paper we present our simulation framework based on ROS and Gazebo targeted to the MORPHEUS rover \cite{giubilato2019evaluation}, a research test-bed for autonomous space operations developed at the University of Padova (see Fig. \ref{fig::morph_real}).}
A replica of the rover (Fig. \ref{fig::morph_replica}) is driven in a variety of simulated planetary environments, enriched with 3D models of rocks of various sizes to add structure.
We evaluate both vision and LiDAR perception technologies. Vision has been extensively used for navigation purposes for NASA MER and MSL rovers, and will be used in the next rover missions: ESA ExoMars and NASA Mars 2020 \cite{Goldberg2002}.
Although to this day no LiDAR sensor have been used on planetary rovers, they have been employed for relative navigation in on-Earth-orbit space missions \cite{christian2013survey}, opening the possibility of future implementation in planetary environments.

This paper is structured as follows: Section \ref{sec::works} introduces recent related works, Section \ref{sec::sim} presents the simulation framework, Section \ref{sec::loc_alg} introduces the tested localization algorithms, Section \ref{sec::exp} reports an in-depth analysis of the results and Section \ref{sec::conc} contains some final remarks.

\section{Related Works}\label{sec::works}
\begin{table*}[!th]
\centering
\caption{Summary of the tested algorithms}\label{tab:algorithms}
\begin{tabular}{|r|c|c|c|}
    \hline
     \textbf{Algorithm} & \textbf{Sensor} & \textbf{Loop Closure} & \textbf{Implementation Notes}\\ \hline\hline
     ORB-SLAM2 \cite{mur2017visual} & Stereo / Mono / RGB-D & \checkmark{} & 2400 maximum ORB features \\
     RTAB-MAP \cite{labbe2019rtab} & Stereo / RGB-D & \checkmark{} & ORB features for Loop Closure - enabled Hypothesis Verification - use g2o \cite{Kummerle2009} \\
     LibVISO2 \cite{Geiger2011IV} & Stereo / Mono & X & - \\ \hline\hline
     A-LOAM \cite{zhang2014loam} & 3D LiDAR & X & - \\
     HDL-SLAM \cite{koide2018portable} & 3D LiDAR & \checkmark{} & scan registration with NDT \cite{biber2003normal} \\
     LeGO-LOAM \cite{legoloam2018} & 3D LiDAR & \checkmark{} & - \\ \hline
\end{tabular}
\end{table*}

In literature exist a variety of research works which make use of the Gazebo simulation environment. Many of them are related to indoor mapping and navigation \cite{10.1007/978-3-319-25903-1_24,wang2017autonomous,rivera2019unmanned}. In \cite{jayasekara2012testing} a minimal simulated environment is used to test the operations of a planetary research platform, and finally building a 3D representation of the observed environment in form of OctoMap \cite{hornung13auro}.

Recently, Gazebo has been used to test and develop navigation strategies for Astrobee \cite{fluckiger2019astrobee,vargas2018astrobee}, a flying robot for the International Space Station. The robot tracks its motion using Visual-Inertial sensing and uses a depth camera to build maps for path planning. All sensors are simulated in a virtual environment replicating the interiors of the ISS, allowing to test the full navigation pipeline in the proper operative conditions.

The authors of \cite{allan2019planetary} used Gazebo to build a simulator for rover operations in a lunar environment. The Gazebo rendering engine has been modified to some extent in order to enable loading of several kilometers wide DTMs while keeping the computational cost at minimum. Photorealism is obtained through visual shaders replicating sun glare, improvements on the shadow generation and by adding custom bump maps to draw wheel marks on the ground.

The authors of \cite{mccrum2010mars}, in order to validate the vision-based algorithm for the ExoMars rover navigation, developed a simulation capable of generating realistic Mars-like images. Their simulation was based on the University   of   Dundee's computer graphics utility PANGU \cite{parkes2009testing}.

\section{Simulated Rover and Test Environment}\label{sec::sim}
The MORPHEUS rover is a mobile platform targeted at unstructured terrains. 6 wheels, individually powered by MAXON\textsuperscript\textregistered motors, are mounted on three rockers passively connected to the rover body by revoluting joints at their barycenter. Turning is performed by skid-steering such that both spot turns and pivot turns are possible. The motor drivers are controlled by Arduino microcontrollers which receive inputs and communicate the motors status to a nVidia Jetson TX2 running Ubuntu 14.04, where all the local processing is done. The Jetson shares a Wi-Fi ROS network with a laptop intented as a base station, where the status of the robot can be monitored and user inputs can be forwarded. The rover is equipped with a Stereolabs ZED camera, which captures synchronized image pairs at variable framerates and resolutions. The stereo processing (distortion correction and stereo rectification) is performed on the embedded Tegra GPU. The rover carries also a plane scanning LiDAR to perform obstacle avoidance.

\subsection{The Rover Model}
An URDF model of the rover is exported from CAD drawings using the ROS add-on for SolidWorks \texttt{sw\_urdf\_exporter}\footnote{wiki.ros.org/sw\_urdf\_exporter}. As the complexity of the model induces a significant computational load to the rendering and physics engine, we provide also a simplified version retaining complete functionality. The skid-steer locomotion is implemented using the \texttt{diff\_drive\_controller}\footnote{wiki.ros.org/diff\_drive\_controller}.

The stereo camera is implemented using the \texttt{multicamera} plugin which allows to simulate lens distortion and noise in the image. We combine this plugin with the recently released \texttt{lens\_flare\_sensor} \cite{allan2019planetary} to simulate the lens flare effect on the image when the sun lies close to the line of sight.
The LiDAR sensor is simulated replicating a Velodyne VLP-16 using the \texttt{gazebo\_ros\_velodyne\_laser}\footnote{wiki.ros.org/velodyne\_gazebo\_plugins} plugin.

\subsection{The Environment}\label{sec::virtual_env}
\renewcommand{\sfdefault}{lmss}
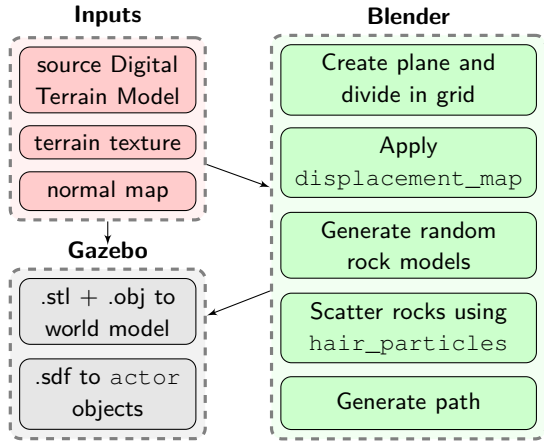
\begin{figure}[t]
	\centering
	\tikzstyle{decision} = [diamond, draw, aspect=1, fill=green!20,
	text width=8em, text badly centered, node distance=3cm, inner sep=2pt]
	\tikzstyle{blockFE} = [rectangle, draw, fill=green!20,
	text width=9em, text centered, rounded corners, minimum height=2em]
	\tikzstyle{blockBE} = [rectangle, draw, fill=cyan!20,
	text width=8em, text centered, rounded corners, minimum height=2em]
	\tikzstyle{empty} = [circle, text width=2.5em, inner sep=0pt, minimum height=1em]
	\tikzstyle{blockC} = [rectangle, draw, fill=black!10,
	text width=6em, text centered, rounded corners, minimum height=2em]
	\tikzstyle{line} = [draw, -latex']
	\tikzstyle{line-short} = [draw, -latex', shorten >=0.5cm]
	\tikzstyle{line-sshort} = [draw, -latex', shorten >=0.3cm]
	\tikzstyle{cloud} = [draw, rectangle, rounded corners,fill=red!20, node distance=2cm,
	minimum height=1em , text centered,
	text width=6em,]
	\begin{tikzpicture}[node distance = 1cm, auto, font=\sf]
	% front end
	\node [cloud, label=below:{}] (dtm) {\small source Digital Terrain Model};
	\node [cloud, below of=dtm, node distance=0.83cm, label=below:{}] (text) {\small terrain texture};
	\node [cloud, below of=text, node distance=0.65cm, label=below:{}] (bmaps) {\small normal map};
	\node [blockFE, right of=dtm, node distance=4cm] (plane) {\small Create plane and divide in grid};
	\node [blockFE, below of=plane, node distance=1.1cm] (dispmap) {\small Apply \texttt{displacement\_map}};
	\node [blockFE, below of=dispmap, node distance=1.1cm] (rocks) {\small Generate random rock models};
	\node [blockFE, below of=rocks, node distance=1.1cm] (scatter) {\small Scatter rocks using \texttt{hair\_particles}};
	\node [blockFE, below of=scatter, node distance=1.0cm] (add1) {\small Generate path};

	% calibration
	\node[blockC, below of=bmaps, node distance=1.6cm] (out) {\small .stl + .obj to world model};
	\node[blockC, below of=out, node distance=1.1cm] (out2) {\small .sdf to \texttt{actor} objects};
	%\node[blockC, below of=calib1, node distance=1.2cm] (calib2) {\small alti correspondence};
	% back end
	% lines
	%\path [line] (image) -- (feat) node[pos=0.5,sloped,above] {$I$};
	%box
	\begin{pgfonlayer}{bg}
	\node[draw=gray,dashed,fill=green!5,line width=1.2pt, rounded corners, label={\small \textbf{Blender}}, fit=(plane) (dispmap) (rocks) (scatter) (add1)](blender) {};
	\end{pgfonlayer}
	%\begin{pgfonlayer}{bg}
	%\node[draw=gray,dashed,fill=cyan!5,line width=1.2pt, rounded corners,  label={\small \textbf{Back-end}}, fit=(blank) (graph) (altic) (isam2)](backend) {};
	%\end{pgfonlayer}
	\begin{pgfonlayer}{bg}
	\node[draw=gray,dashed,fill=red!5,line width=1.2pt, rounded corners,  label={\small \textbf{Inputs}}, fit=(dtm) (text) (bmaps) ](Inputs) {};
	\end{pgfonlayer}
	\begin{pgfonlayer}{bg}
	\node[draw=gray,dashed,fill=black!5,line width=1.2pt, rounded corners,  label={\small \textbf{Gazebo}}, fit=(out) (out2)](outs) {};
	\end{pgfonlayer}
	%\path [line] (calib) -- (backend);
	\path [line] (Inputs) -- (blender);
	\path [line] (blender) -- (outs);
	\path [line-sshort] (Inputs) -- (outs);
	%\path [line-short] (sensors) -- (calib) node[pos=0.4,right] {$\rho$};
	%\path [line] (pose) -- (outFE);
	%\path [line-sshort] (blank) -- (pose) node[pos=0.4,left] {\small $\{\mathbf{X}_j^W, \mathbf{T}_k^W\}$};
	\end{tikzpicture}
	\caption{Schematic workflow to generate a virtual environment in Gazebo from a Digital Terrain Model using the open source 3D modeler Blender}
	\label{fig::terrain::overview}
\end{figure}
\renewcommand{\sfdefault}{phv}

The virtual environment is modeled after a Digital Terrain Model (DTM) of the Gale crater on Mars, cropped to a planar landscape. A schematic overview of the map generation process is given in Fig. \ref{fig::terrain::overview}.
The DTM is imported in Gazebo to create a base featureless surface as a basis for the virtual environment. To populate the surface with rocks, we import the DTM in the 3D modeler Blender applying a displacement map to a plane segmented in a coarse grid.
Two population of rocks are scattered over this surface using a manually weighted random distribution roughly matching the frequencies observed on the Martian surface \cite{golombek2003rock}:
a small population of large boulders and a
large population of smaller rocks with diameters ranging from 0.1 to 0.5 meters.
We released the environments in form of a ROS package\footnote{github.com/MorpheusPD/MarsSim}.

To precisely evaluate the performances of SLAM algorithms on this environment, we use Blender to generate two fixed paths along which the robot will move using the \texttt{actor} functionalities of ROS Gazebo.
%\scmargin{}{Forse scriverei qualcosa di piu' generico. Anche il fatto di esportare il file CSV è un dettaglio che non serve metterei}
%Paths are created modifying \texttt{NURBS} Circle objects to suitable closed shapes and finally applying
%a \texttt{Shrinkwrap} modifier to project them on the terrain plus an offset.
To simulate the motion caused by the roughness of the terrain we added noise to the camera orientations.
The resulting sequences of poses are exported to
%CSV files, for easier handling in Matlab, and to
SDF files to instruct the Gazebo \texttt{actor} objects.
%A Follow Path Object Constraint was applied to the Camera object and using the Graph Editor the motion profile of the rover was designed.
%Some rotation angle noise was added to the Camera Object to simulate the motion caused by the rover navigating on a rocky terrain.
%A custom blender/python script was developed to export the \texttt{bpy.context.scene.objects} data containing the pose information on a CSV file for every twelth of a second of the rover motion on the path.
%Another script was developed to export the pose information to a SDFormat file containing the Gazebo Actor.

\section{Localization Algorithms}\label{sec::loc_alg}
In this paper we compare a variety of odometry or SLAM algorithms using either the virtual stereo camera or 3D LiDAR to provide localization (i.e. compute the transformation between the robot and world reference frames $\mathbf{T}_\text{robot}^{w}$) showing how our virtual environment can be used to aid the design choices for the perception system of a robot depending on the target environment. An overview is provided in Table \ref{tab:algorithms} along with relevant implementation remarks about parameter values that differ from the default ones.

\subsection{Visual SLAM}
Among all available Visual SLAM algorithms for stereo cameras, we selected ORB-SLAM2 \cite{mur2017visual}, RTAB-MAP \cite{labbe2019rtab} and LibVISO2 \cite{Geiger2011IV}. LiBVISO2 is a widely used Visual Odometry algorithm without Loop Closure capabilities. RTAB-MAP is instead a RGB-D Graph SLAM with a bayesian Loop Closure detector that addresses multi-session mapping and is highly configurable through an easy Graphical User Interface (GUI) (e.g. types of feature detectors and descriptors, optimizers and respective parameters). ORB-SLAM2 is a Visual SLAM algorithm for monocular, stereo and RGB-D vision systems based on ORB features \cite{rublee2011orb} which leverages a Bag of Words approach \cite{Methods::GalvezTRO12} for localization and Loop Closure detection.

\subsection{LiDAR SLAM}
In addition, we compare the performances of a variety of recently published LiDAR SLAM algorithms which are released open source and are compatible with the Robot Operating System. A-LOAM\footnote{github.com/HKUST-Aerial-Robotics/A-LOAM} is an implementation of LOAM \cite{Method::LOAM} where odometry and mapping are decoupled to be performed at a faster and slower rate respectively and the poses are computed by matching edge and planar features across scans. LeGO-LOAM \cite{legoloam2018} improves the performances of LOAM by extracting and matching point clusters across LiDAR scans and by explicitly utilizing the ground to constrain the roll pitch and z coordinates during pose tracking. In addition to the original LOAM, a pose graph is maintained to include Loop Closures. Finally we test \texttt{hdl\_graph\_slam}\footnote{github.com/koide3/hdl\_graph\_slam} \cite{koide2018portable} (referred here as HDL-SLAM), an open source LiDAR SLAM package for the Robot Operating System which provides a modular graph SLAM for 3D LiDARs based on scan registration through ICP or NDT \cite{NDT}. It provides interfaces for easy integration of IMU and GPS measurements and performs Loop Closure detection to correct a pose graph.

\section{Experiments and Discussion}\label{sec::exp}
\begin{table}[t]
    \centering
    \caption{Root Mean Square of Absolute Trajectory Error and Median of Translation Drift in the \textit{Long} Sequence}
    \begin{tabular}{|r|c|c|c|c|c|c|}
        \hline
         & \textbf{ORB} & \textbf{RTAB} & \textbf{VISO2} & \textbf{ALOAM} & \textbf{HDL} & \textbf{LeGO} \\
         \hline\hline
         ATE [m]  &  0.48 & \textbf{0.14} & 1.56 & 0.76 & 34.29 & 0.21 \\
         TDr [\%] & 6.31 & 0.43 & \textbf{0.34} & 0.81 & 60.3 & 0.47 \\
     \hline
    \end{tabular}
    \label{tab:rmse_long}
\end{table}
\begin{table}[t]
    \centering
    \caption{Root Mean Square of Absolute Trajectory Error and Median of Translation Drift in the \textit{Short} Sequence}
    \begin{tabular}{|r|c|c|c|c|c|c|}
        \hline
         & \textbf{ORB} & \textbf{RTAB} & \textbf{VISO2} & \textbf{ALOAM} & \textbf{HDL} & \textbf{LeGO} \\
         \hline\hline
         ATE [m]  &  0.07 & \textbf{0.04} & 0.39 & 7.65 & 2.49 & 0.63 \\
         TDr [\%] & 1.91 & 0.62 & \textbf{0.55} & 7.58 & 13.1 & 2.89 \\
     \hline
    \end{tabular}
    \label{tab:rmse_short}
\end{table}
%In this paper we evaluate a variety of localization approaches in the simulated environment. Specifically we test the Visual SLAM algorithm ORB-SLAM2 \cite{Methods::ORB2} against A-LOAM\footnote{github.com/HKUST-Aerial-Robotics/A-LOAM}, a LiDAR-only SLAM adapted from LOAM \cite{Method::LOAM}.
%In this section we report the results of the experiments performed with both SLAM algorithms (see an example in Fig. \ref{fig::eval}).
We performed a variety of experiments in two sequences generated as explained in Sec. \ref{sec::virtual_env}.
The first sequence, denominated \textit{Long}, takes places in the environment visible in Fig.\ref{fig::gale} which comprises a denser distribution of small pebbles and a sparser distribution or larger rocks with dimensions comparable to the ones of the rover. A closed trajectory, about 300 meters long, allows to evaluate the tracking performances of all algorithms in the presence of 90$^\circ$ turns as well as the Loop Closure capabilities, as the rovers returns in the initial location with the same viewpoint.
A second and shorter sequence, denominated here \textit{Short}, is about 60 meters long and takes place around high boulders, generally bigger than the rover. An approximately triangular trajectory ends in proximity of the beginning, however on an opposite camera viewpoint, not allowing detection of Loop Closures from the visual pipelines but, in principle, allowing it for LiDAR pipelines which benefit from 360$^\circ$ range coverage.

%Concerning the rover perception the camera is a ZED stereo camera (image size -- $1270\times720$; FOV -- $110^{\circ}$) and the LiDAR is a Velodyne VLP-16 (horizontal FOV -- $360^{\circ}$; vertical FOV -- $30^{\circ}$; 16 scanning planes).
%The metrological performances of the tested algorithm have been evaluated in term of Absolute Trajectory Error (ATE).
The virtual rover is equipped with a stereo camera whose specifications make it equivalent to a Stereolabs ZED stereo camera, which is mounted on our Morpheus rover (see Fig.~{\ref{fig::morph_real}}). The 3D LiDAR is modeled roughly after the Ouster OS-1 with 64 scan planes. The full characteristics of both sensors are reported for brevity in Table~\ref{tab:sensors_spec}.

\begin{table}[ht]
    \centering
    \caption{Camera and LiDAR characteristics}
    \begin{tabular}{|r|c|c|}
    \hline
                    & \textbf{Stereo camera} & \textbf{3D LiDAR} \\ \hline\hline
         Resolution & 1280x720 px & 0.2$^\circ$ (H) x ~0.4$^\circ$ (V)\\
         FoV & 90° (H) x 60° (V) & 90° (H) x 30° (V) \\
         Refresh Rate & 30 Hz & 10 Hz \\
         Baseline & 0.12 m& - \\
         \hline
    \end{tabular}
    \label{tab:sensors_spec}
\end{table}
%Figure \ref{fig::trajectories} shows the trajectories  as  obtained  during  the  simulation  and  reconstructed by  the  SLAM  algorithms.
%The ATE error of the proposed methods is shown in \figurename{\ref{fig::ATE_error}}.
%It is possible to observe that in the simulated configuration A-LOAM gives better performances than ORB-SLAM, this because A-LOAM benefits from the 360-degree scan of the lidar, while ORB-SLAM suffers from the lack of features when the rover passes the rocky area.
We test the performances of the algorithms introduces in Sec.~\ref{sec::loc_alg} in terms of how accurately they reconstruct the trajectories from the \textit{Long} and \textit{Short} sessions. We first align the trajectory to the ground truth using Horn's method \cite{Horn} given pose correspondences found by matching timestamps. In order to not underestimate the pose errors resulting from angular drift, only the first third of the whole trajectory is use for alignment. For each correspondence, we compute the Absolute Trajectory Error (ATE), or the L2 distance between the aligned poses:
\begin{equation}
    \text{ATE}_i = ||\mathbf{x}_i^* - \mathbf{x}_i||_\text{L2}
\end{equation}
where $\mathbf{x}_i^*$ and $\mathbf{x}_i$ are positions of corresponding poses from ground truth and estimated from SLAM respectively after alignment. We also compute the translation drift as the relative difference between the lengths of local segments of the estimated trajectory and ground truth. Let be $\mathbf{x}_i^*$ and $\mathbf{x}_j^*$ two ground truth poses such that the length of the trajectory that connects them $l(\mathbf{x}_i^*, \mathbf{x}_j^*)$ is 10 meters. Let then be $\mathbf{x}_i$ and $\mathbf{x}_j$ the estimated poses from SLAM that correspond via timestamps to $\mathbf{x}_i^*$ and $\mathbf{x}_j^*$. Thus, we define the local translation drift as:
\begin{equation}
    \text{TDr}_i = \frac{|l(\mathbf{x}_i^*, \mathbf{x}_j^*) - l(\mathbf{x}_i, \mathbf{x}_j)|}{l(\mathbf{x}_i^*, \mathbf{x}_j^*)}
\end{equation}
Finally, we report a summary of the results in a table for both sequences by computing the Root Mean Square of the errors for each time point along the trajectories and for each algorithm.
\begin{figure}[tb]
    \centering
        \subfloat[Visual SLAM map]{%
           \includegraphics[width=\linewidth, trim=0 0 0 0cm, clip]{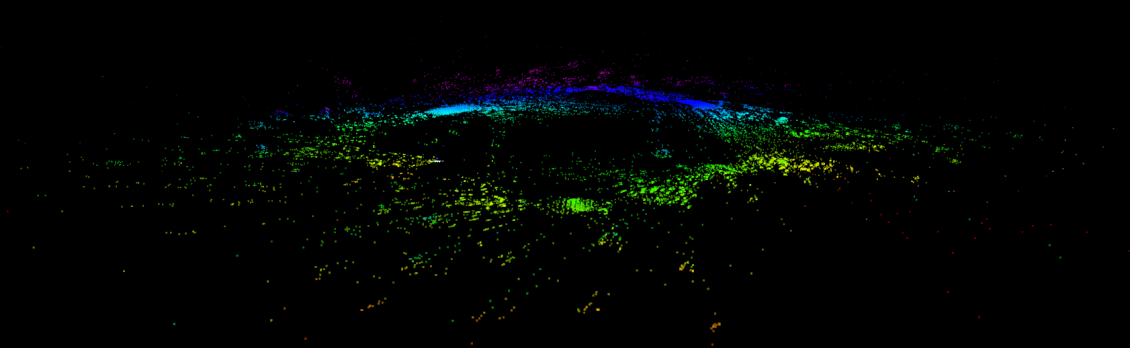}\label{fig::visual_eval}} \\
        \subfloat[LiDAR SLAM map]{%
           \includegraphics[width=\linewidth]{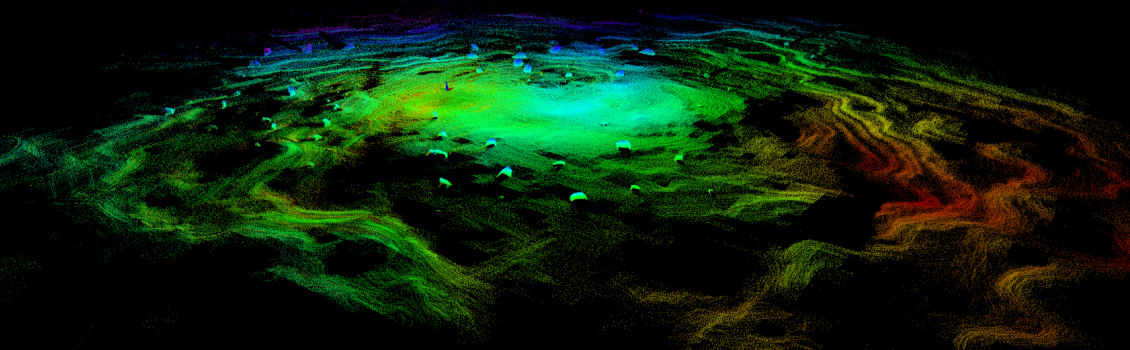}\label{fig::lidar_eval}}
    \caption{Visualization of the maps built from a visual SLAM (ORB-SLAM2) and a LiDAR SLAM (LeGO-LOAM). This figures highlight the different appearance from a sparse visual map of 3D landmarks from detected image features and a dense LiDAR map obtained by stacking 3D LiDAR scans given accurate estimations of the sensor poses}
    \label{fig::map_comparison}
\end{figure}
%\begin{figure*}[!ht]
%    \subfloat{\includesvg[width=0.45\linewidth]{long/leg}}  \hfill
%    \subfloat{\includesvg[width=0.45\linewidth]{long/leg}} \\
%    \subfloat[Trajectories \& environment \textbf{\textit{Long} sequence}]{\includesvg[width=0.45\linewidth]{long/traj_long2}\label{fig:long:traj}}  \hfill
%    \subfloat[Trajectories \& environment \textbf{\textit{Short} sequence}]{\includesvg[width=0.47\linewidth]{short/traj_short2}\label{fig:short:traj}} \\
%    \subfloat[Absolute Trajectory Error]{\includesvg[width=0.45\linewidth]{long/ate_long}\label{fig:long:ate}} \hfill
%    \subfloat[Absolute Trajectory Error]{\includesvg[width=0.45\linewidth]{short/ate_short}\label{fig:short:ate}} \\
%    \subfloat[Translation Drift]{\includesvg[width=0.45\linewidth]{long/drift_long}\label{fig:long:dr}} \hfill
%    \subfloat[Translation Drift]{\includesvg[width=0.45\linewidth]{short/drift_short}\label{fig:short:dr}}
%    \caption{Performances and result visualization of the compared stereo and LiDAR SLAM systems in the \textit{Long} and \textit{Short} sequences. Trajectories are overlaid on top views of the environment, showing the amount and distribution of rocks. The ATE plots focus only on the algorithms that succeeded in estimating the trajectory.}
%    \label{fig:algo_performances}
%\end{figure*}
\begin{figure*}[!ht]
    \subfloat{\def\svgwidth{0.45\linewidth}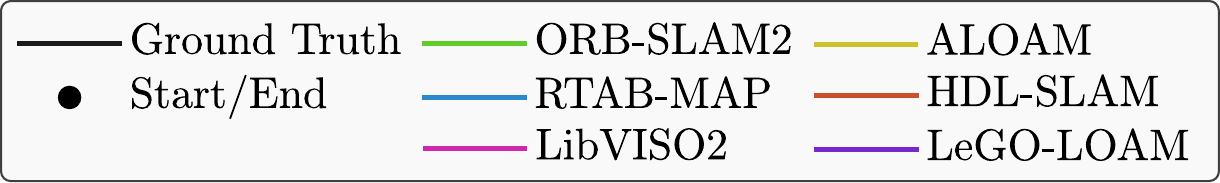}  \hfill
    \subfloat{\def\svgwidth{0.45\linewidth}\input{svg-inkscape/leg_svg-tex.pdf_tex}} \\
    \subfloat[Trajectories \& environment \textbf{\textit{Long} sequence}]{\def\svgwidth{0.45\linewidth}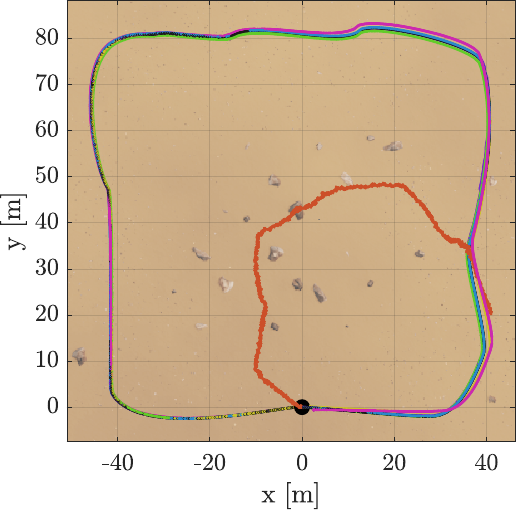\label{fig:long:traj}}  \hfill
    \subfloat[Trajectories \& environment \textbf{\textit{Short} sequence}]{\def\svgwidth{0.45\linewidth}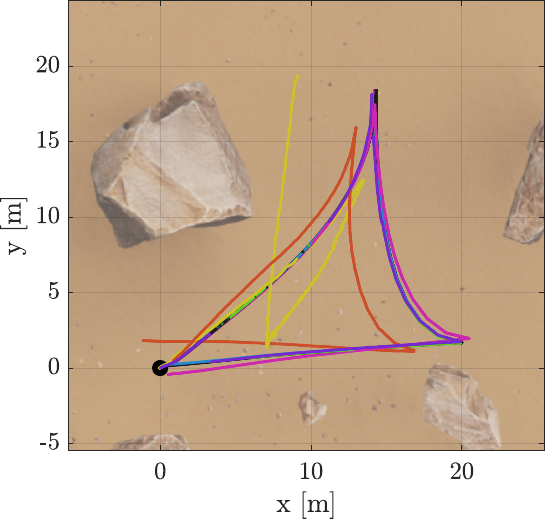\label{fig:short:traj}} \\
    \subfloat[Absolute Trajectory Error]{\def\svgwidth{0.45\linewidth}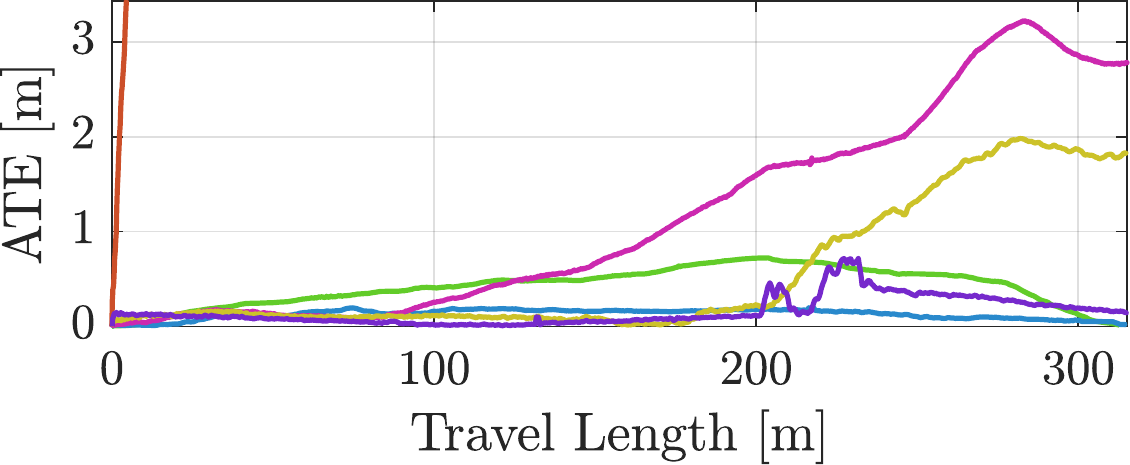\label{fig:long:ate}} \hfill
    \subfloat[Absolute Trajectory Error]{\def\svgwidth{0.45\linewidth}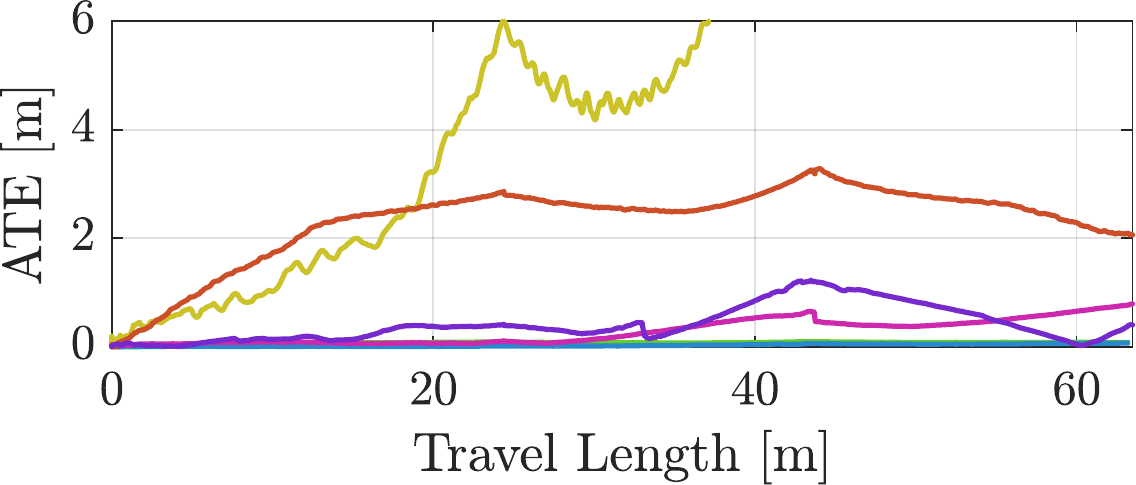\label{fig:short:ate}} \\
    \subfloat[Translation Drift]{\def\svgwidth{0.45\linewidth}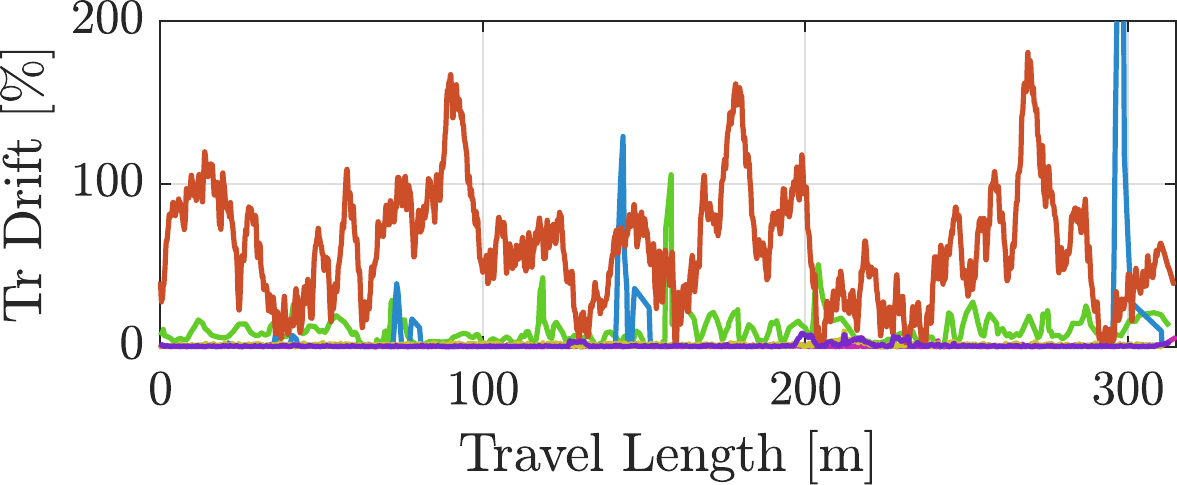\label{fig:long:dr}} \hfill
    \subfloat[Translation Drift]{\def\svgwidth{0.45\linewidth}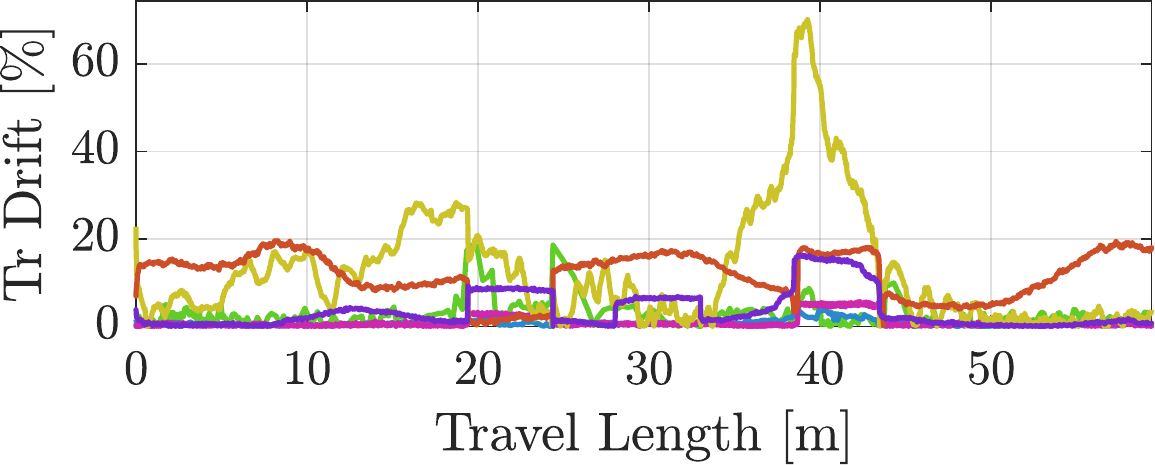\label{fig:short:dr}}
    \caption{Performances and result visualization of the compared stereo and LiDAR SLAM systems in the \textit{Long} and \textit{Short} sequences. Trajectories are overlaid on top views of the environment, showing the amount and distribution of rocks. The ATE plots focus only on the algorithms that succeeded in estimating the trajectory.}
    \label{fig:algo_performances}
\end{figure*}

A first mean of comparison among visual and LiDAR approaches is in the quality and density of the map, which might be employed for the detection of geological features. Figure \ref{fig::map_comparison} presents the qualitative difference between a visual map, comprised of sparse 3D landmarks, and a LiDAR map, built by concatenating LiDAR scans. A quantitative evaluation of performances is instead presented in
Figure \ref{fig:algo_performances}, which reports the results of all algorithms on the \textit{Long} and \textit{Short} sequences. In addition, Tables \ref{tab:rmse_long} and \ref{tab:rmse_short} contain the RMS errors highlighting the best scoring algorithms. Figures \ref{fig:long:traj} and \ref{fig:short:traj} show the resulting trajectories and ground truth overlaid on a top-view of the environment to highlight the context in terms of geometry.
In the \textit{Long} sequence both ORB-SLAM2 and RTAB-MAP were able to successfully close the loop, therefore their ATE is close to zero at both the beginning and end of the trajectory. However, ORB-SLAM2 accumulates some translation drift which manifest itself as higher ATEs in the middle of the trajectory (see Fig. \ref{fig:long:ate}). Contrarily, LibVISO2 exhibits the lowest translational drift but accumulates angular drift which can not be recovered as it is a pure visual odometry. RTAB-MAP instead shows consistent performances in both sequences achieving the lowest ATE errors thanks to an accurate odometry and Loop Closure capabilities. The LiDAR odometry A-LOAM outperform the visual odometry LibVISO2 in the \textit{Long} sequence although obtaining the highest errors in the \textit{Short} sequence. LeGO-LOAM instead even outperforms ORB-SLAM in the \textit{Long} sequence.
This result is surprising given the little geometric structures present in this sequence but can be explained given that both ALOAM and LeGO-LOAM extract and match edge features belonging to the LiDAR scans, of which the environment has plenty and uniformly distributed, contrarily to the \textit{Short} sequence which is characterized by bigger and sparsely distributed boulders that obstruct the view. HDL-SLAM instead relies on the mechanism of scan matching, which degenerates in the presence on planar scenes. This explains the extreme translation drift in the \textit{Long} sequence.

\section{Conclusions}\label{sec::conc}
In this work we presented a simulation framework for mobile robots based on ROS Gazebo. We demonstrated how it can be used to aid the selection of perception sensors based on the expected geometry and appearance of the environment. Furthermore, we compared the performances of a variety of open source Visual and LiDAR SLAM algorithms in different environments characterized by different rock distributions and size. Although visual SLAM proves to be accurate in presence of textured ground, LiDAR SLAM has the advantage of building detailed maps in form of point clouds. As for future developments of this work, we plan to enhance the photo-realism of the simulation and join the advantages of both SLAM approaches, fusing 3D LiDARs with stereo cameras using the simulated environment to validate the approach.

\section*{Acknowledgement}
This work has been supported by Project ``ARES", Progetti Innovativi degli Studenti, University of Padova. We would also like to thank the Morpheus Team for the discussions and participation in the experiments.

\bibliographystyle{IEEEtran}
\bibliography{biblio}

\end{document}